\documentclass[sigconf,authorversion]{acmart}

\AtBeginDocument{%
  \providecommand\BibTeX{{%
    \normalfont B\kern-0.5em{\scshape i\kern-0.25em b}\kern-0.8em\TeX}}}

\setcopyright{rightsretained}
\copyrightyear{2022}
\acmYear{2022}
\acmDOI{XXXXXXX.XXXXXXX}

\acmConference[WWW22]{TheWebConf Workshop on Graph Learning Benchmarks 2022}{April 26, 2022}{Virtual}

\usepackage{acro}
\usepackage[acronym, nomain, nonumberlist]{glossaries}
\DeclareAcronym{MR}{
    short = MR,
    long  = mean rank
}
\DeclareAcronym{ZMR}{
    short = ZMR,
    long  = z-mean rank
}
\DeclareAcronym{MRR}{
    short = MRR,
    long  = mean reciprocal rank
}
\DeclareAcronym{ZMRR}{
    short = ZMRR,
    long  = z-mean reciprocal rank
}
\DeclareAcronym{HK}{
    short = H\textsubscript{$k$},
    long  = hits at $k$
}
\DeclareAcronym{ZHK}{
    short = ZH\textsubscript{$k$},
    long  = z-hits at $k$
}
\DeclareAcronym{AHK}{
    short = AH\textsubscript{$k$},
    long  = adjusted hits at $k$
}
\DeclareAcronym{AMRR}{
    short = AMRR,
    long  = adjusted mean reciprocal rank
}
\DeclareAcronym{AMR}{
    short = AMR,
    long  = adjusted mean rank
}
\DeclareAcronym{AMRI}{
    short = AMRI,
    long  = adjusted mean rank index
}
\DeclareAcronym{HMR}{
    short = HMR,
    long  = harmonic mean rank
}
\DeclareAcronym{GMR}{
    short = GMR,
    long  = geometric mean rank
}
\DeclareAcronym{IGMR}{
    short = IGMR,
    long  = inverse geometric mean rank
}
\DeclareAcronym{IMR}{
    short = IMR,
    long  = inverse mean rank
}
\DeclareAcronym{KGEM}{
    short = KGEM,
    long  = knowledge graph embedding model
}
\DeclareAcronym{KG}{
    short = KG,
    long  = knowledge graph
}
\DeclareAcronym{CWA}{
    short = CWA,
    long  = closed-world assumption
}
\DeclareAcronym{OWA}{
    short = OWA,
    long  = open-world assumption
}

\usepackage{xcolor,pifont}

\definecolor{cycle2}{RGB}{106, 191, 0}
\definecolor{cycle3}{RGB}{191, 0, 0}

\newcommand{\cmark}{\textcolor{cycle2}{\ding{52}}} %
\newcommand{\xmark}{\textcolor{cycle3}{\ding{56}}}

\newcommand{\Var}{\mathbb{V}\mathrm{ar}}

\begin{document}

\title[Ranking Metrics]{A Unified Framework for Rank-based Evaluation Metrics for Link Prediction in Knowledge Graphs}

\author{Charles Tapley Hoyt}
\authornote{Both authors contributed equally to this research.}
\affiliation{%
  \institution{Harvard Medical School}
  \city{Boston}
  \state{MA}
  \country{USA}
}
\email{cthoyt@gmail.com}
\orcid{0000-0003-4423-4370}

\author{Max Berrendorf}
\authornotemark[1]
\affiliation{%
  \institution{LMU Munich}
  \city{M\"unchen}
  \country{Germany}}
\email{max.berrendorf@gmail.com}
\orcid{0000-0001-9724-4009}

\author{Mikhail Galkin}
\affiliation{%
  \institution{Mila \& McGill University}
  \city{Montreal}
  \state{QC}
  \country{Canada}
}
\email{mikhail.galkin@mila.quebec}
\orcid{0000-0003-3526-0155}

\author{Volker Tresp}
\affiliation{%
  \institution{LMU Munich \& Siemens AG}
  \city{M\"unchen}
  \country{Germany}}
\email{volker.tresp@siemens.com}
\orcid{0000-0001-9428-3686}

\author{Benjamin M. Gyori}
\affiliation{%
  \institution{Harvard Medical School}
  \city{Boston}
  \state{MA}
  \country{USA}
}
\email{benjamin_gyori@hms.harvard.edu}
\orcid{0000-0001-9439-5346}

\renewcommand{\shortauthors}{Hoyt and Berrendorf, \textit{et al.}}

\begin{abstract}
    The link prediction task on knowledge graphs without explicit negative triples in the training data motivates the usage of rank-based metrics.
    Here, we review existing rank-based metrics and propose desiderata for improved metrics to address lack of interpretability and comparability of existing metrics to datasets of different sizes and properties.
    We introduce a simple theoretical framework for rank-based metrics upon which we investigate two avenues for improvements to existing metrics via alternative aggregation functions and concepts from probability theory.
    We finally propose several new rank-based metrics that are more easily interpreted and compared accompanied by a demonstration of their usage in a benchmarking of knowledge graph embedding models.
\end{abstract}



\maketitle

\section{Introduction}

\Acp{KG} are a structured formalism for representing facts about entities $\mathcal{E}$ and their relationships $\mathcal{R}$ as triples of the form $(h,r,t) \in \mathcal{E} \times \mathcal{R} \times \mathcal{E}$. \acp{KG} are useful for entity clustering, link prediction, entity disambiguation, question answering, dialogue systems, and recommendation systems~\cite{wang2017}.
They can be constructed under one of two assumptions: under the \ac{CWA}, the non-existence of a triple in the \ac{KG} implies its falsiness and under the \ac{OWA}, the non-existence of a triple in the \ac{KG} neither implies its falsiness nor truthiness.
Most real-world \acp{KG} are constructed under the \ac{OWA} to reflect their relative incompleteness with respect to true triples and typical lack of triples known to be false.

Link prediction on \acp{KG} constructed under the \ac{OWA} is a popular approach for addressing their relative incompleteness that can be conceptualized as a binary classification task on triples.
However, the lack of false triples leads to a positive unlabeled learning scenario~\cite{bekker2020pu} which requires negative sampling i.e., randomly labeling some unknown triples as false during the training and evaluation of machine learning models like \acp{KGEM}.
Partially because these techniques introduce bias to classification metrics whose formulation depends on true negatives and false negatives (such as the accuracy, $F_1$, and ROC-AUC), the last ten years of \ac{KGEM} literature has nearly exclusively used the rank-based evaluation metrics: \ac{HK}, \ac{MR}, and \ac{MRR}.

Despite their ubiquity, these metrics are not comparable when applied to datasets of different sizes and properties and lack a corresponding theoretical framework for describing their properties and shortcomings.
For example, there is recent interest in applying link prediction with \acp{KGEM} to real-world tasks in biomedicine such as drug repositioning, target identification, and side effect prediction.
However, there are several choices of biomedical \acp{KG} for training and evaluation, each formulated with different entities, relations, and source databases~\cite{bonner2021}.
Because the choice of \ac{KG} can meaningfully impact downstream applications, hyperparameter search and evaluation must also consider and compare different \acp{KG}, which, without rank-based evaluation metrics that are comparable across datasets nor theory to help articulate the issue, is not possible.

This work addresses these limitations by making the following contributions:
\textbf{(1)} Proposes a theoretical foundation for rank-based evaluation metrics; \textbf{(2)} Proposes and characterizes novel rank-based evaluation metrics with alternative rank transformations and alternative aggregation operations based on special cases of the generalized power mean~\cite{Bullen2003}; \textbf{(3)} Derives probabilistic adjustments for existing and novel rank-based evaluation metrics inspired by~\cite{berrendorf2020,tiwari2021}.
We implemented the proposed novel metrics and made them available as part of the PyKEEN~\cite{ali2021} software package\footnote{\url{https://github.com/pykeen/pykeen}}.

\section{Background}

\subsection{Generating Ranks}

Given a set of entities $\mathcal{E}$, a set of relations $\mathcal{R}$, a \acl{KG} $\mathcal{K} \subseteq \mathcal{E} \times \mathcal{R} \times \mathcal{E}$, disjoint testing and validation triples $\mathcal{T}_{test}, \mathcal{T}_{eval} \subseteq \mathcal{K}$, and a \ac{KGEM} with scoring function $g:\mathcal{E} \times \mathcal{R} \times \mathcal{E} \rightarrow \mathbb{R}$ (e.g., TransE~\cite{bordes2013} has $g(h,r,t)=-\|\mathbf{e}_h+\mathbf{r}_r-\mathbf{e}_t\|_2$), evaluation concurrently solves two tasks as described by~\cite{nickel2015review}:

\begin{enumerate}
    \item In right-side prediction, for each triple $(h, r, t) \in \mathcal{T}_{eval}$, we score candidate triples $\{(h, r, e) \mid e \in \mathcal{E}\}$ using $g$.
    \item In left-side prediction, for each triple $(h, r, t) \in \mathcal{T}_{eval}$, we score candidate triples $\{(e, r, t) \mid e \in \mathcal{E}\}$ using $g$.
\end{enumerate}

In each task, candidate triples are sorted in descending order based on their scores, and are then assigned a rank based on their 1-indexed sort position.
In the (optional) filtered setting proposed by~\cite{bordes2013}, candidate triples appearing in $\mathcal{T}_{test}$ are removed from the ordering so they do not artificially increase the ranks of true triples ($\mathcal{T}_{eval}$) appearing later in the sorted list.
A good model results in low ranks $r_1,\dots,r_n$ for the true triples, reflecting its ability to assign high scores to true triples and low scores to negatively sampled triples.
The ranks are typically aggregated using a rank-based evaluation metric to quantify the performance of the \ac{KGEM} with a single number.

While the upper bound on an individual rank $r_i$ is generally $|\mathcal{E}|$ for both the right- and left-side prediction tasks, the number of candidates may be (considerably) smaller than $|\mathcal{E}|$, e.g., in the filtered setting~\cite{bordes2013} or during sampled evaluation~\cite{teru2020,hu2021}\footnote{E.g., for OGB-LSC WikiKG2, 1,001 candidates are considered for $|\mathcal{E}|\approx80M$, i.e., $n < 10^{-5}|\mathcal{E}|$.}.


\subsection{Aggregating Ranks}\label{sec:metrics-survey}

While the distribution of raw ranks gives full insight into evaluation performance, it is much more convenient to report aggregate statistics.
In this subsection, we introduce and describe three common rank-based metrics reported in applications and evaluation of link prediction: \acl{HK}, \acl{MR}, and \acl{MRR}. 

\paragraph{\Acl{HK}}

The \acf{HK} (\autoref{eq:hk-definition}) is an increasing metric (i.e., higher values are better) that captures the fraction of true entities that appear in the first $k$ entities of the sorted rank list.
Thus, it is tailored towards a use-case where only the top-$k$ entries are to be considered, e.g., due to a limited number of results shown on a search result page.
Because it does not differentiate between cases when the rank is larger than $k$, a miss with rank $k+1$ and $k+d$ where $d \gg 1$ have the same effect on the final score.
Therefore, it is less suitable for the comparison of different models.
\begin{equation} \label{eq:hk-definition}
\textrm{H}_k(r_1,\ldots,r_n) = \frac{1}{n}\sum_{i=1}^n  \mathbb{I}[r_i \leq k] \hspace{5mm} \in [0, 1]
\end{equation}
where the indicator function $\mathbb{I}$ is defined as $\mathbb{I}[x \leq y] = 1$ for $x \leq y$ and $0$ otherwise.


\paragraph{\Acl{MR}}

The \acf{MR} (\autoref{eq:mr-definition}) is a decreasing metric (i.e., lower values are better) that corresponds to the arithmetic mean over ranks of true triples.
It has the advantage over \ac{HK} that it is non-parametric and better reflects average performance.

\begin{equation} \label{eq:mr-definition}
\textrm{MR}(r_1,\ldots,r_n) = \frac{1}{n}\sum_{i=1}^n r_i \hspace{5mm} \in [1,\infty]
\end{equation}

\paragraph{\Acl{MRR}}

The \acf{MRR} (\autoref{eq:mrr-definition}) is an increasing metric that corresponds to the arithmetic mean over the reciprocals of ranks of true triples.
The construction of the \ac{MRR} biases it towards changes in low ranks without completely disregarding high ranks like the \ac{HK}.
It can therefore be considered as a soft a version of \ac{HK} that is less sensitive to outliers and is often used during early stopping due to this behavior.
While it has been argued that \ac{MRR} has theoretical flaws \cite{fuhr2018}, these arguments are not undisputed~\cite{sakai2020}.

\begin{equation} \label{eq:mrr-definition}
\textrm{MRR}(r_1,\ldots,r_n) = \frac{1}{n}\sum_{i=1}^n r_i^{-1} \hspace{5mm} \in (0,1]
\end{equation}

\subsection{Desiderata}

\begin{table}
    \small
    \centering
    \def\arraystretch{1.2}
    \caption{%
    Desiderata for rank-based metrics
    }
    \label{tab:desiderata}
    \begin{tabular}{ l l c c c} 
    \toprule
    \textbf{Property}   & \textbf{Constraint}  & \textbf{\ac{MR}} & \textbf{\ac{MRR}} & \textbf{\ac{HK}}                   \\ 
    \midrule
    Non-negativity      & $\forall r\in \mathbb{N}: f(r) \geq 0$ & \cmark & \cmark & \cmark \\
    Fixed optimum       & $f(1) = c_{\text{opt.}}$                             & \xmark  & \cmark  & \cmark \\
    Asymp. pessimum & $\lim_{r \to \infty} f(r) = c_{\text{pes.}}$         & \xmark  & \cmark & \cmark \\
    Anti-monotonic   & $r > r' \rightarrow f(r) < f(r')$   & \xmark  & \cmark & \xmark  \\
    Size invariant      & $\mathbb{E}[f] \not\propto n$                                      & \xmark  & \xmark  & \xmark  \\
    \bottomrule
    \end{tabular}
\end{table}
After examining the strengths and weaknesses of the three most common rank-based metrics, we outline five desiderata for rank-based metrics that are interpretable and comparable across models/datasets/evaluation tasks in \autoref{tab:desiderata}.
Because each of \ac{MR}, \ac{MRR}, and \ac{HK} can be defined with the strictly monotonic increasing arithmetic mean as the aggregation function, we describe our desiderata with respect to the transformation $f$ applied to $r_i$ within the aggregation (further explored in \autoref{sec:rank-transformations}).

We first propose the intuitive constraint that the co-domain of the metric is non-negative, which is satisfied by all three metrics.
We next propose that the best rank should result in a fixed optimum (ideally, 1) and that worse ranks should asymptotically approach a fixed pessimum (ideally, 0 for unbalanced metrics and -1 for balanced metrics), which are both satisfied by both \ac{MRR} and \ac{HK} but neither by \ac{MR}.

We propose that along this gradient, the metric should be strictly anti-monotonic, meaning that,  as the performance of the \ac{KGEM} improves, the predictions for true triples should improve, result in lower ranks $r_i$, increased values of the transformed ranks $f(r_i)$, and increased evaluation metrics.
This is only satisfied by \ac{MRR} as \ac{MR} is an increasing function (i.e., higher ranks result in higher scores) and $f(r_i)$ for \ac{HK} has only two discrete values (0 and 1), and thus is not strict in terms of monotonicity.

Because the worst rank is bounded based on the number of candidate triples (which itself depends on the dataset and the evaluation procedure) rather than a constant, the same \ac{MR} from the evaluation on two different datasets of different sizes should not be directly compared.
While \ac{MRR} and \ac{HK} are bounded with a constant, the shape of these curves are affected by the same properties of the number of candidate triples and the same issue is applicable.
This situation makes the interpretation of results from even large robustness and ablation studies (whose aims are to identify patterns in interaction models, loss functions, regularizations, and other properties of KGEM) challenging at best and misleading at worst.
Therefore, our final desideratum is that rank-based metrics should be invariant to the number of candidate triples and directly comparable across different datasets and evaluation procedures.

\section{An Analysis of Rank-based Metrics}

A general form of a rank-based metric $\mathbb{M}$ over ranks $r_1,\dots,r_n \in \mathbb{N}$ is $\mathbb{M}(r_1,\ldots,r_n) = g\big(\oplus_{i=1}^n f(r_i)\big)$ where $f:\mathbb{N} \mapsto \mathbb{R}$ is a rank transformation function, $\oplus: \mathbb{R}^n \mapsto \mathbb{R}$ is an aggregation operation, and $g:\mathbb R \mapsto \mathbb{R}$ is a post-aggregation transformation function.

\subsection{Insight and Discovery \textit{via} Transformations}\label{sec:rank-transformations}

The \ac{MR}, \ac{MRR}, and \ac{HK} metrics use the arithmetic mean as an aggregation function with $\oplus_{i=1}^n f(r_i) = \frac{1}{n}\sum_{i=1}^n f(r_i)$ and varying definitions for rank transformation function $f$ and post-aggregation function $g$ as summarized in \autoref{tab:transformations}.
Based on this formulation, we propose that in addition to the colloquial formulation of \ac{MRR} using the arithmetic mean and $f(x)=x^{-1}$, it can additionally be formulated with the harmonic mean and $g(x)=x^{-1}$.
We suggest a more descriptive name for this metric could be the inverse harmonic mean rank, or IHMR.
Considering the definition with the harmonic mean best explains why \ac{MRR} has the desired properties of an asymptotic pessimum that the \ac{MR} lacks.
It further motivates the construction of two counterpart metrics, the \ac{IMR} and the \ac{HMR} (respective inverses of \ac{MR} and \ac{MRR}) which are included in \autoref{tab:transformations}.

\begin{table}
    \small
    \centering
    \caption{%
        The identity, reciprocal, and discrete indicator functions are used in combination with various aggregation functions to define the \ac{MR}, \ac{MRR}, \ac{HK}, and four novel metrics.
        The aggregation (agg.) column uses the $M_p$ notation of the generalized Hölder mean described in Appendix \autoref{tab:aggregations}.
    }
    \label{tab:transformations}
    \def\arraystretch{1.2}
    \begin{tabular}{ l c l c} 
    \toprule
    \textbf{Metric}  & $f(x)$ & \textbf{Agg.} &  $g(x)$ \\ 
    \midrule
    \acs{HK}         & $\mathbb{I}[x \leq k]$ & $M_1$    & $ x$ \\
    \acs{MR}         & $x$                      & $M_1$    & $ x$ \\
    \acs{MRR} & $x$                      & $M_{-1}$ & $ x^{-1}$ \\
    \acs{MRR} (colloquial)       & $x^{-1}$                 & $M_1$    & $ x$ \\
    \midrule
    \acs{IMR}      & $x$  & $M_1$ & $ x^{-1}$ \\ 
    \acs{HMR}        & $x$  & $M_{-1}$ & $ x$ \\ 
    \acs{GMR}        & $x$  & $M_0$ & $ x$ \\ 
    \acs{IGMR}        & $x$  & $M_0$ & $ x^{-1}$ \\ 
    \bottomrule
    \end{tabular}
\end{table}

\subsection{Insight and Discovery \textit{via} Aggregations}

While the typical aggregation of ranks applied after various transformations is the arithmetic mean, $\frac{1}{n}\sum_{i=1}^n f(r_i)$, here we present alternate aggregations in \autoref{tab:aggregations}.
The max rank and min rank would not be useful in practice due to their susceptibility to outliers, but they nicely demonstrate the bounds on the three Pythagorean means and the quadratic means.
%
Through the lens of the Pythagorean means (i.e., special cases of the generalized Hölder mean in \autoref{tab:aggregations}), we can better explain why \ac{MR} (and therefore also \ac{IMR}) tends to bias towards high ranks and \ac{MRR} (and therefore also \ac{HMR}) tends to bias towards low ranks.
The aggregation that compromises best between the arithmetic mean and harmonic mean is the geometric mean, therefore, we use it to define the \ac{GMR} and \ac{IGMR}.

\section{Probabilistic Adjustments}

Inspired by the probabilistic adjustments to \ac{MR} that resulted in the \acf{AMR} and the \acf{AMRI}~\cite{berrendorf2020}, we considered generalizing their derivations and applying them to \ac{MRR} and \ac{HK}.
Similar to~\cite{berrendorf2020}, we assume the ranking tasks $i$ to be independent, and the ranks uniformly discretely distributed over $[1, \ldots, N_i]$, such that $r_i \sim \mathcal{U}(1, N_i)$.
Note that $N_i$ may not be constant across ranking tasks $i$ due to filtered evaluation~\cite{bordes2013}.

\subsection{Adjustments}

\textbf{Expectation Adjustment.}
The derivation of the \ac{AMR} motivated normalizing a base metric $\mathbb{M}$ by its expectation such that $\mathbb{M}^{*}(r_1,\ldots,r_n)=\frac{\mathbb{M}(r_1,\ldots,r_n)}{\mathbb{E}\left[\mathbb{M}\right]}$.
We found that it was only useful for metrics bounded by $[1, \infty)$ (i.e.,\ \ac{MR}, \ac{HMR}, \ac{GMR}) whose adjustments were bounded by $[0, 1)$ and not for metrics bounded by $(0, 1]$ (i.e.,\ \ac{IMR}, \ac{MRR}, \ac{IGMR}) whose adjustments were unbounded on $(0,\infty)$.
The expected value of the adjusted metric is thus $1$.
Because it is not generally applicable, we do not propose any new metrics using this adjustment.

\begin{figure*}[t]
    \centering
    \includegraphics[width=\linewidth]{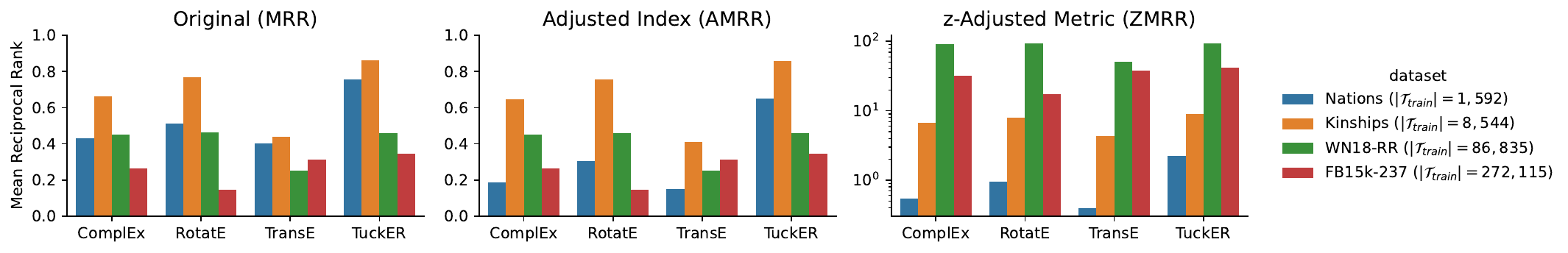}
    \caption{Original, adjusted index, and $z$-adjusted metric for the \acf{MRR} (inverse harmonic mean rank). Datasets are presented in increasing size from left to right Nations having the least and FB15k-237 having the most.}
    \label{fig:benchmarking}
\end{figure*}

\textbf{Adjusted Index.}
The derivation of the \ac{AMRI} motivated normalizing a base metric $\mathbb{M}$ by its expectation then linearly transforming it such that the maximum value maps to 1, the expectation maps to 0, positive values can be considered good, and negative values can be considered bad such that $
\mathbb{M}^{*}(r_1,\ldots,r_n) = \frac{\mathbb{M}(r_1,\ldots,r_n) - \mathbb{E}\left[\mathbb{M}\right]}{\max(\mathbb{M}) - \mathbb{E}\left[\mathbb{M}\right]}$.
We use this form to propose the \ac{AHK} and \ac{AMRR}.

Surprisingly, the derivation of \ac{AMRI} from \ac{MR} resulted in the same form $\frac{\mathbb{M} - \mathbb{E}[\mathbb{M}]}{1 - \mathbb{E}[\mathbb{M}]}$ for the \ac{AMRR} and the \ac{AHK}, despite their different monotonicities (i.e., increasing or decreasing) and co-domains.
We note that these characteristics do result in different lower bound behavior, which for \ac{AMRI} is a constant $-1$ and for the \ac{AMRR} and \ac{AHK} is a function of the expectation of the base metric $-\frac{\mathbb{E}[\mathbb{M}]}{1-\mathbb{E}[\mathbb{M}]}$.
Related derivations can be found in \autoref{sec:adjustements}.


\textbf{$z$-Adjustment.}
We finally propose a novel probabilistic adjustment enabled by the central limit theorem.
Because \ac{MR}, \ac{MRR}, \ac{HK}, and other metrics $\mathbb{M}$ are defined as the sum of random variables (despite their several transformations), they have asymptotic Gaussian characteristics.
Therefore, we propose using the standarization technique $z=\frac{x-\mu}{\sigma}$ to define a z-scored metric $
\mathbb{M}^{*}(r_1,\ldots,r_n) = \frac{\mathbb{M}(r_1,\ldots,r_n) - \mathbb{E}\left[\mathbb{M}\right]}{\sqrt{\Var\left[\mathbb{M}\right]}}$.
We apply this to the \ac{MR}, \ac{MRR}, and \ac{HK} to respectively define three new metrics: \ac{ZMR}, \ac{ZMRR}, and \ac{ZHK}.
Related derivations can be found in \autoref{sec:adjustements}.

We note that the z-scored metrics can be monotonically mapped with the cumulative distribution function of the standard normal distribution onto the interval $(0,1)$ to fulfill the desiderata.
However, we suggest that z-scores are adequately interpretable and comparable without transformation, as the number of standard deviations below or above of the expected value.

\subsection{Discussion}
Each of the three probabilistic adjustment strategies presented in this work are affine transformations of the base metric with scale and bias constants only dependent on the studied ranking task, but independent of the investigated predictions.
Thus, they can be applied to the base metrics after computation of a pre-computed expectation and variance that are appropriate for the dataset, e.g., to make results from existing publications more comparable across datasets and splits.
We provide a database of pre-computed expectations and variances for benchmark datasets included in PyKEEN~\cite{ali2021} stratified by split (i.e., training, testing, validation), evaluation task (left-hand, right-hand, both), and metric on Zenodo~\cite{hoyt2022adjustments_data}.

\subsection{Case Study}\label{sec:case-study}

In order to demonstrate the improved interpretability and comparability of our newly proposed adjustmented metrics and z-scored metrics, we re-evaluated four \acp{KGEM} (ComplEx~\cite{DBLP:journals/jmlr/TrouillonDGWRB17}, RotatE~\cite{sun2018rotate}, TransE~\cite{bordes2013}, TuckER~\cite{DBLP:conf/emnlp/BalazevicAH19}) on four datasets (WN18-RR~\cite{DBLP:conf/aaai/DettmersMS018}, FB15k-237~\cite{toutanova2015}, Nations, and Kinships~\cite{Kemp2006}) of varying size (from 14 entities to 40k entities, see Appendix \autoref{tab:datasets}) reusing the optimal hyperparameters reported in~\cite{ali2020benchmark}.

\autoref{fig:benchmarking} presents a comparison between original metric \ac{MRR}, its adjusted index \ac{AMRR}, and its $z$-adjustment \ac{ZMRR}.
We first observe that the \ac{MRR} displays an anti-correlation with size of each dataset that is not present for \ac{AMRR} and \ac{ZMRR}, disregarding the smallest dataset for which the numerical behavior of the adjustments is slightly erratic.
While the original metric suggests that ComplEx performs similarly on WN18-RR (green) and Nations (blue), the adjusted metric shows that the difference is more remarkable.
Conversely, the original metric suggests that TuckER performs better on Nations than WN18-RR, while the adjusted metric shows that when improving comparability by adjusting for size effects, TuckER actually performs better on WN18-RR.

Finally, the $z$-adjusted metric enables direct comparison between the results on different datasets while also giving insight into their significance by normalizing against the expectation and variance of the metric under random rankings.
This adjustment reveals that the improved original metrics on the two smaller datasets (Kinships and Nations) were less significant than the results on the two larger datasets (WN18-RR and FB15k-237), despite achieving better unnormalized performance.

All configuration, trained models, results, and analysis presented in this case study are available at \url{https://github.com/pykeen/ranking-metrics-manuscript} and archived on Zenodo at~\cite{hoyt2022adjustments_repo}.

\section{Conclusion}

In this article, we motivated and reviewed rank-based evaluation metrics for the link prediction task on \acp{KG} before proposing desiderata for metrics with improved interpretability and comparability.
We developed a simple theoretical framework for describing rank-based evaluation metrics, investigated their probabilistic properties, and ultimately proposed several new metrics with desired properties based on alternate aggregation functions (i.e.,\ \ac{HMR}, \ac{GMR}), alternate transformations (i.e., \ac{IMR},\ \ac{IGMR}), and probabilistic adjustments (i.e.,\ \ac{AMRR}, \ac{AHK}, \ac{ZMR}, \ac{ZMRR}, and \ac{ZHK}).
We provide implementations of these metrics in PyKEEN~\cite{ali2021} v1.8.0 with closed form solutions for the expectation and variance of the base metrics when possible and numeric solutions for the rest.
We leave the remaining derivations of closed forms for the metrics defined with more complicated functions (e.g., \ac{GMR}) for future work,  to enable generation of z-scored metrics for the remaining base metrics.

\textit{Generalization.}
While we restricted our description in this work to the evaluation of link prediction on \acp{KG}, the discussed approaches are directly applicable to other settings which use rank-based evaluation, e.g., the entity pair ranking protocol~\cite{DBLP:conf/rep4nlp/WangRGBM19}, entity alignment~\cite{DBLP:conf/ijcai/WuLF0Y019,DBLP:conf/iclr/FeyL0MK20,DBLP:journals/pvldb/SunZHWCAL20,DBLP:conf/cikm/MaoWXWL20}, query embedding~\cite{hamilton2018embedding,ren2020query2box,ren2020beta,DBLP:conf/aaai/KotnisLN21,arakelyan2021complex,alivanistos2022query}, uni-relational link prediction~\cite{DBLP:conf/nips/HuFZDRLCL20,DBLP:journals/corr/abs-2106-06935}, and relation detection~\cite{DBLP:conf/icpr/SharifzadehBBKT20}.

\textit{Future Work.}
Existing evaluation frameworks commonly compute one rank value per evaluation triple and side, then aggregate the ranks.
However, real-world \acp{KG} often contain hub entities that occur in many triples which may therefore dominate the evaluation.
We intend to build on previous work~\cite{DBLP:conf/icpr/SharifzadehBBKT20,tiwari2021,alivanistos2022query}, investigating the issue using our novel metrics in a deeper investigation of rank-based evaluation of the link prediction task on \aclp{KG}.


\begin{acks}

Charles Tapley Hoyt and Benjamin M. Gyori were supported by the DARPA Young Faculty Award W911NF20102551.
Max Berrendorf was supported by the German Federal Ministry of Education and Research (BMBF) under Grant No. 01IS18036A.
Mikhail Galkin was supported by the Samsung AI grant held at Mila.

\end{acks}
\clearpage
\bibliographystyle{ACM-Reference-Format}
\bibliography{main}


\begin{thebibliography}{36}


\ifx \showCODEN    \undefined \def \showCODEN     #1{\unskip}     \fi
\ifx \showDOI      \undefined \def \showDOI       #1{#1}\fi
\ifx \showISBNx    \undefined \def \showISBNx     #1{\unskip}     \fi
\ifx \showISBNxiii \undefined \def \showISBNxiii  #1{\unskip}     \fi
\ifx \showISSN     \undefined \def \showISSN      #1{\unskip}     \fi
\ifx \showLCCN     \undefined \def \showLCCN      #1{\unskip}     \fi
\ifx \shownote     \undefined \def \shownote      #1{#1}          \fi
\ifx \showarticletitle \undefined \def \showarticletitle #1{#1}   \fi
\ifx \showURL      \undefined \def \showURL       {\relax}        \fi
\providecommand\bibfield[2]{#2}
\providecommand\bibinfo[2]{#2}
\providecommand\natexlab[1]{#1}
\providecommand\showeprint[2][]{arXiv:#2}

\bibitem[Ali et~al\mbox{.}(2020)]%
        {ali2020benchmark}
\bibfield{author}{\bibinfo{person}{Mehdi Ali}, \bibinfo{person}{Max
  Berrendorf}, \bibinfo{person}{Charles~Tapley Hoyt}, \bibinfo{person}{Laurent
  Vermue}, \bibinfo{person}{Mikhail Galkin}, \bibinfo{person}{Sahand
  Sharifzadeh}, \bibinfo{person}{Asja Fischer}, \bibinfo{person}{Volker Tresp},
  {and} \bibinfo{person}{Jens Lehmann}.} \bibinfo{year}{2020}\natexlab{}.
\newblock \showarticletitle{{Bringing Light Into the Dark: A Large-scale
  Evaluation of Knowledge Graph Embedding Models Under a Unified Framework}}.
\newblock \bibinfo{journal}{\emph{arXiv}} (\bibinfo{date}{jun}
  \bibinfo{year}{2020}).
\newblock
\showeprint[arxiv]{2006.13365}
\urldef\tempurl%
\url{http://arxiv.org/abs/2006.13365}
\showURL{%
\tempurl}


\bibitem[Ali et~al\mbox{.}(2021)]%
        {ali2021}
\bibfield{author}{\bibinfo{person}{Mehdi Ali}, \bibinfo{person}{Max
  Berrendorf}, \bibinfo{person}{Charles~Tapley Hoyt}, \bibinfo{person}{Laurent
  Vermue}, \bibinfo{person}{Sahand Sharifzadeh}, \bibinfo{person}{Volker
  Tresp}, {and} \bibinfo{person}{Jens Lehmann}.}
  \bibinfo{year}{2021}\natexlab{}.
\newblock \showarticletitle{{PyKEEN 1.0: A Python Library for Training and
  Evaluating Knowledge Graph Embeddings}}.
\newblock \bibinfo{journal}{\emph{Journal of Machine Learning Research}}
  \bibinfo{volume}{22}, \bibinfo{number}{82} (\bibinfo{year}{2021}),
  \bibinfo{pages}{1--6}.
\newblock
\urldef\tempurl%
\url{http://jmlr.org/papers/v22/20-825.html}
\showURL{%
\tempurl}


\bibitem[Alivanistos et~al\mbox{.}(2022)]%
        {alivanistos2022query}
\bibfield{author}{\bibinfo{person}{Dimitrios Alivanistos}, \bibinfo{person}{Max
  Berrendorf}, \bibinfo{person}{Michael Cochez}, {and} \bibinfo{person}{Mikhail
  Galkin}.} \bibinfo{year}{2022}\natexlab{}.
\newblock \showarticletitle{Query Embedding on Hyper-Relational Knowledge
  Graphs}. In \bibinfo{booktitle}{\emph{International Conference on Learning
  Representations}}.
\newblock
\urldef\tempurl%
\url{https://openreview.net/forum?id=4rLw09TgRw9}
\showURL{%
\tempurl}


\bibitem[Arakelyan et~al\mbox{.}(2021)]%
        {arakelyan2021complex}
\bibfield{author}{\bibinfo{person}{Erik Arakelyan}, \bibinfo{person}{Daniel
  Daza}, \bibinfo{person}{Pasquale Minervini}, {and} \bibinfo{person}{Michael
  Cochez}.} \bibinfo{year}{2021}\natexlab{}.
\newblock \showarticletitle{Complex Query Answering with Neural Link
  Predictors}. In \bibinfo{booktitle}{\emph{International Conference on
  Learning Representations}}.
\newblock
\urldef\tempurl%
\url{https://openreview.net/forum?id=Mos9F9kDwkz}
\showURL{%
\tempurl}


\bibitem[Balazevic et~al\mbox{.}(2019)]%
        {DBLP:conf/emnlp/BalazevicAH19}
\bibfield{author}{\bibinfo{person}{Ivana Balazevic}, \bibinfo{person}{Carl
  Allen}, {and} \bibinfo{person}{Timothy~M. Hospedales}.}
  \bibinfo{year}{2019}\natexlab{}.
\newblock \showarticletitle{TuckER: Tensor Factorization for Knowledge Graph
  Completion}. In \bibinfo{booktitle}{\emph{Proceedings of the 2019 Conference
  on Empirical Methods in Natural Language Processing and the 9th International
  Joint Conference on Natural Language Processing, {EMNLP-IJCNLP} 2019, Hong
  Kong, China, November 3-7, 2019}}, \bibfield{editor}{\bibinfo{person}{Kentaro
  Inui}, \bibinfo{person}{Jing Jiang}, \bibinfo{person}{Vincent Ng}, {and}
  \bibinfo{person}{Xiaojun Wan}} (Eds.). \bibinfo{publisher}{Association for
  Computational Linguistics}, \bibinfo{pages}{5184--5193}.
\newblock


\bibitem[Bekker and Davis(2020)]%
        {bekker2020pu}
\bibfield{author}{\bibinfo{person}{Jessa Bekker} {and} \bibinfo{person}{Jesse
  Davis}.} \bibinfo{year}{2020}\natexlab{}.
\newblock \showarticletitle{{Learning from positive and unlabeled data: a
  survey}}.
\newblock \bibinfo{journal}{\emph{Mach. Learn.}} \bibinfo{volume}{109},
  \bibinfo{number}{4} (\bibinfo{year}{2020}), \bibinfo{pages}{719--760}.
\newblock
\showISSN{1573-0565}
\urldef\tempurl%
\url{https://doi.org/10.1007/s10994-020-05877-5}
\showDOI{\tempurl}


\bibitem[Berrendorf et~al\mbox{.}(2020)]%
        {berrendorf2020}
\bibfield{author}{\bibinfo{person}{Max Berrendorf}, \bibinfo{person}{Evgeniy
  Faerman}, \bibinfo{person}{Laurent Vermue}, {and} \bibinfo{person}{Volker
  Tresp}.} \bibinfo{year}{2020}\natexlab{}.
\newblock \showarticletitle{{On the Ambiguity of Rank-Based Evaluation of
  Entity Alignment or Link Prediction Methods}}.
\newblock \bibinfo{journal}{\emph{arXiv}} (\bibinfo{date}{feb}
  \bibinfo{year}{2020}).
\newblock
\showeprint[arxiv]{2002.06914}
\urldef\tempurl%
\url{http://arxiv.org/abs/2002.06914}
\showURL{%
\tempurl}


\bibitem[Bonner et~al\mbox{.}(2021)]%
        {bonner2021}
\bibfield{author}{\bibinfo{person}{Stephen Bonner}, \bibinfo{person}{Ian~P
  Barrett}, \bibinfo{person}{Cheng Ye}, \bibinfo{person}{Rowan Swiers},
  \bibinfo{person}{Ola Engkvist}, \bibinfo{person}{Andreas Bender},
  \bibinfo{person}{Charles~Tapley Hoyt}, {and} \bibinfo{person}{William
  Hamilton}.} \bibinfo{year}{2021}\natexlab{}.
\newblock \showarticletitle{{A Review of Biomedical Datasets Relating to Drug
  Discovery: A Knowledge Graph Perspective}}.
\newblock  (\bibinfo{date}{feb} \bibinfo{year}{2021}).
\newblock
\showeprint[arxiv]{2102.10062}
\urldef\tempurl%
\url{http://arxiv.org/abs/2102.10062}
\showURL{%
\tempurl}


\bibitem[Bordes et~al\mbox{.}(2013)]%
        {bordes2013}
\bibfield{author}{\bibinfo{person}{Antoine Bordes}, \bibinfo{person}{Nicolas
  Usunier}, \bibinfo{person}{Alberto Garcia-Dur\'{a}n}, \bibinfo{person}{Jason
  Weston}, {and} \bibinfo{person}{Oksana Yakhnenko}.}
  \bibinfo{year}{2013}\natexlab{}.
\newblock \showarticletitle{Translating Embeddings for Modeling
  Multi-Relational Data}. In \bibinfo{booktitle}{\emph{Proceedings of the 26th
  International Conference on Neural Information Processing Systems - Volume
  2}} (Lake Tahoe, Nevada) \emph{(\bibinfo{series}{NIPS'13})}.
  \bibinfo{publisher}{Curran Associates Inc.}, \bibinfo{address}{Red Hook, NY,
  USA}, \bibinfo{pages}{2787–2795}.
\newblock


\bibitem[Bullen(2003)]%
        {Bullen2003}
\bibfield{author}{\bibinfo{person}{P.~S. Bullen}.}
  \bibinfo{year}{2003}\natexlab{}.
\newblock \bibinfo{booktitle}{\emph{{Handbook of Means and Their
  Inequalities}}}.
\newblock \bibinfo{publisher}{Springer Netherlands},
  \bibinfo{address}{Dordrecht}.
\newblock
\showISBNx{978-90-481-6383-0}
\urldef\tempurl%
\url{https://doi.org/10.1007/978-94-017-0399-4}
\showDOI{\tempurl}


\bibitem[Dettmers et~al\mbox{.}(2018)]%
        {DBLP:conf/aaai/DettmersMS018}
\bibfield{author}{\bibinfo{person}{Tim Dettmers}, \bibinfo{person}{Pasquale
  Minervini}, \bibinfo{person}{Pontus Stenetorp}, {and}
  \bibinfo{person}{Sebastian Riedel}.} \bibinfo{year}{2018}\natexlab{}.
\newblock \showarticletitle{Convolutional 2D Knowledge Graph Embeddings}. In
  \bibinfo{booktitle}{\emph{Proceedings of the Thirty-Second {AAAI} Conference
  on Artificial Intelligence, (AAAI-18)}}. \bibinfo{publisher}{{AAAI} Press},
  \bibinfo{pages}{1811--1818}.
\newblock


\bibitem[Fey et~al\mbox{.}(2020)]%
        {DBLP:conf/iclr/FeyL0MK20}
\bibfield{author}{\bibinfo{person}{Matthias Fey}, \bibinfo{person}{Jan~Eric
  Lenssen}, \bibinfo{person}{Christopher Morris}, \bibinfo{person}{Jonathan
  Masci}, {and} \bibinfo{person}{Nils~M. Kriege}.}
  \bibinfo{year}{2020}\natexlab{}.
\newblock \showarticletitle{Deep Graph Matching Consensus}. In
  \bibinfo{booktitle}{\emph{8th International Conference on Learning
  Representations, {ICLR} 2020, Addis Ababa, Ethiopia, April 26-30, 2020}}.
  \bibinfo{publisher}{OpenReview.net}.
\newblock
\urldef\tempurl%
\url{https://openreview.net/forum?id=HyeJf1HKvS}
\showURL{%
\tempurl}


\bibitem[Fuhr(2018)]%
        {fuhr2018}
\bibfield{author}{\bibinfo{person}{Norbert Fuhr}.}
  \bibinfo{year}{2018}\natexlab{}.
\newblock \showarticletitle{{Some Common Mistakes In IR Evaluation, And How
  They Can Be Avoided}}.
\newblock \bibinfo{journal}{\emph{SIGIR Forum}} \bibinfo{volume}{51},
  \bibinfo{number}{3} (\bibinfo{date}{feb} \bibinfo{year}{2018}),
  \bibinfo{pages}{32--41}.
\newblock
\showISSN{0163-5840}
\urldef\tempurl%
\url{https://doi.org/10.1145/3190580.3190586}
\showDOI{\tempurl}


\bibitem[Hamilton et~al\mbox{.}(2018)]%
        {hamilton2018embedding}
\bibfield{author}{\bibinfo{person}{William~L. Hamilton}, \bibinfo{person}{Payal
  Bajaj}, \bibinfo{person}{Marinka Zitnik}, \bibinfo{person}{Dan Jurafsky},
  {and} \bibinfo{person}{Jure Leskovec}.} \bibinfo{year}{2018}\natexlab{}.
\newblock \showarticletitle{Embedding Logical Queries on Knowledge Graphs}. In
  \bibinfo{booktitle}{\emph{Advances in Neural Information Processing Systems
  31: Annual Conference on Neural Information Processing Systems 2018, NeurIPS
  2018, December 3-8, 2018, Montr{\'{e}}al, Canada}}.
  \bibinfo{pages}{2030--2041}.
\newblock


\bibitem[Hoyt and Berrendorf(2022)]%
        {hoyt2022adjustments_data}
\bibfield{author}{\bibinfo{person}{Charles~Tapley Hoyt} {and}
  \bibinfo{person}{Max Berrendorf}.} \bibinfo{year}{2022}\natexlab{}.
\newblock \bibinfo{booktitle}{\emph{Rank-based Metric Adjustments}}.
\newblock
\urldef\tempurl%
\url{https://doi.org/10.5281/zenodo.6331629}
\showDOI{\tempurl}


\bibitem[Hoyt et~al\mbox{.}(2022)]%
        {hoyt2022adjustments_repo}
\bibfield{author}{\bibinfo{person}{Charles~Tapley Hoyt}, \bibinfo{person}{Max
  Berrendorf}, {and} \bibinfo{person}{Michael Galkin}.}
  \bibinfo{year}{2022}\natexlab{}.
\newblock \bibinfo{booktitle}{\emph{pykeen/ranking-metrics-manuscript:}}.
\newblock
\urldef\tempurl%
\url{https://doi.org/10.5281/zenodo.6347429}
\showDOI{\tempurl}


\bibitem[Hu et~al\mbox{.}(2021)]%
        {hu2021}
\bibfield{author}{\bibinfo{person}{Weihua Hu}, \bibinfo{person}{Matthias Fey},
  \bibinfo{person}{Hongyu Ren}, \bibinfo{person}{Maho Nakata},
  \bibinfo{person}{Yuxiao Dong}, {and} \bibinfo{person}{Jure Leskovec}.}
  \bibinfo{year}{2021}\natexlab{}.
\newblock \showarticletitle{{OGB-LSC:} {A} Large-Scale Challenge for Machine
  Learning on Graphs}.
\newblock \bibinfo{journal}{\emph{CoRR}}  \bibinfo{volume}{abs/2103.09430}
  (\bibinfo{year}{2021}).
\newblock
\showeprint[arXiv]{2103.09430}
\urldef\tempurl%
\url{https://arxiv.org/abs/2103.09430}
\showURL{%
\tempurl}


\bibitem[Hu et~al\mbox{.}(2020)]%
        {DBLP:conf/nips/HuFZDRLCL20}
\bibfield{author}{\bibinfo{person}{Weihua Hu}, \bibinfo{person}{Matthias Fey},
  \bibinfo{person}{Marinka Zitnik}, \bibinfo{person}{Yuxiao Dong},
  \bibinfo{person}{Hongyu Ren}, \bibinfo{person}{Bowen Liu},
  \bibinfo{person}{Michele Catasta}, {and} \bibinfo{person}{Jure Leskovec}.}
  \bibinfo{year}{2020}\natexlab{}.
\newblock \showarticletitle{Open Graph Benchmark: Datasets for Machine Learning
  on Graphs}. In \bibinfo{booktitle}{\emph{Advances in Neural Information
  Processing Systems 33: Annual Conference on Neural Information Processing
  Systems 2020, NeurIPS 2020, December 6-12, 2020, virtual}}.
\newblock


\bibitem[Kemp et~al\mbox{.}(2006)]%
        {Kemp2006}
\bibfield{author}{\bibinfo{person}{Charles Kemp}, \bibinfo{person}{Joshua~B.
  Tenenbaum}, \bibinfo{person}{Thomas~L. Griffiths}, \bibinfo{person}{Takeshi
  Yamada}, {and} \bibinfo{person}{Naonori Ueda}.}
  \bibinfo{year}{2006}\natexlab{}.
\newblock \showarticletitle{{Learning systems of concepts with an infinite
  relational model}}.
\newblock \bibinfo{journal}{\emph{Proceedings of the National Conference on
  Artificial Intelligence}}  \bibinfo{volume}{1} (\bibinfo{year}{2006}),
  \bibinfo{pages}{381--388}.
\newblock
\showISBNx{1577352815}


\bibitem[Kotnis et~al\mbox{.}(2021)]%
        {DBLP:conf/aaai/KotnisLN21}
\bibfield{author}{\bibinfo{person}{Bhushan Kotnis}, \bibinfo{person}{Carolin
  Lawrence}, {and} \bibinfo{person}{Mathias Niepert}.}
  \bibinfo{year}{2021}\natexlab{}.
\newblock \showarticletitle{Answering Complex Queries in Knowledge Graphs with
  Bidirectional Sequence Encoders}. In \bibinfo{booktitle}{\emph{Thirty-Fifth
  {AAAI} Conference on Artificial Intelligence, {AAAI} 2021, Thirty-Third
  Conference on Innovative Applications of Artificial Intelligence, {IAAI}
  2021, The Eleventh Symposium on Educational Advances in Artificial
  Intelligence, {EAAI} 2021, Virtual Event, February 2-9, 2021}}.
  \bibinfo{publisher}{{AAAI} Press}, \bibinfo{pages}{4968--4977}.
\newblock
\urldef\tempurl%
\url{https://ojs.aaai.org/index.php/AAAI/article/view/16630}
\showURL{%
\tempurl}


\bibitem[Mao et~al\mbox{.}(2020)]%
        {DBLP:conf/cikm/MaoWXWL20}
\bibfield{author}{\bibinfo{person}{Xin Mao}, \bibinfo{person}{Wenting Wang},
  \bibinfo{person}{Huimin Xu}, \bibinfo{person}{Yuanbin Wu}, {and}
  \bibinfo{person}{Man Lan}.} \bibinfo{year}{2020}\natexlab{}.
\newblock \showarticletitle{Relational Reflection Entity Alignment}. In
  \bibinfo{booktitle}{\emph{{CIKM} '20: The 29th {ACM} International Conference
  on Information and Knowledge Management, Virtual Event, Ireland, October
  19-23, 2020}}, \bibfield{editor}{\bibinfo{person}{Mathieu d'Aquin},
  \bibinfo{person}{Stefan Dietze}, \bibinfo{person}{Claudia Hauff},
  \bibinfo{person}{Edward Curry}, {and} \bibinfo{person}{Philippe
  Cudr{\'{e}}{-}Mauroux}} (Eds.). \bibinfo{publisher}{{ACM}},
  \bibinfo{pages}{1095--1104}.
\newblock
\urldef\tempurl%
\url{https://doi.org/10.1145/3340531.3412001}
\showDOI{\tempurl}


\bibitem[Nickel et~al\mbox{.}(2015)]%
        {nickel2015review}
\bibfield{author}{\bibinfo{person}{Maximilian Nickel}, \bibinfo{person}{Kevin
  Murphy}, \bibinfo{person}{Volker Tresp}, {and} \bibinfo{person}{Evgeniy
  Gabrilovich}.} \bibinfo{year}{2015}\natexlab{}.
\newblock \showarticletitle{A review of relational machine learning for
  knowledge graphs}.
\newblock \bibinfo{journal}{\emph{Proc. IEEE}} \bibinfo{volume}{104},
  \bibinfo{number}{1} (\bibinfo{year}{2015}), \bibinfo{pages}{11--33}.
\newblock


\bibitem[Ren et~al\mbox{.}(2020)]%
        {ren2020query2box}
\bibfield{author}{\bibinfo{person}{Hongyu Ren}, \bibinfo{person}{Weihua Hu},
  {and} \bibinfo{person}{Jure Leskovec}.} \bibinfo{year}{2020}\natexlab{}.
\newblock \showarticletitle{Query2box: Reasoning over Knowledge Graphs in
  Vector Space Using Box Embeddings}. In \bibinfo{booktitle}{\emph{8th
  International Conference on Learning Representations, {ICLR} 2020, Addis
  Ababa, Ethiopia, April 26-30, 2020}}. \bibinfo{publisher}{OpenReview.net}.
\newblock
\urldef\tempurl%
\url{https://openreview.net/forum?id=BJgr4kSFDS}
\showURL{%
\tempurl}


\bibitem[Ren and Leskovec(2020)]%
        {ren2020beta}
\bibfield{author}{\bibinfo{person}{Hongyu Ren} {and} \bibinfo{person}{Jure
  Leskovec}.} \bibinfo{year}{2020}\natexlab{}.
\newblock \showarticletitle{Beta Embeddings for Multi-Hop Logical Reasoning in
  Knowledge Graphs}. In \bibinfo{booktitle}{\emph{Advances in Neural
  Information Processing Systems 33: Annual Conference on Neural Information
  Processing Systems 2020, NeurIPS 2020, December 6-12, 2020, virtual}},
  \bibfield{editor}{\bibinfo{person}{Hugo Larochelle},
  \bibinfo{person}{Marc'Aurelio Ranzato}, \bibinfo{person}{Raia Hadsell},
  \bibinfo{person}{Maria{-}Florina Balcan}, {and} \bibinfo{person}{Hsuan{-}Tien
  Lin}} (Eds.).
\newblock


\bibitem[Sakai(2020)]%
        {sakai2020}
\bibfield{author}{\bibinfo{person}{Tetsuya Sakai}.}
  \bibinfo{year}{2020}\natexlab{}.
\newblock \showarticletitle{{On Fuhr's guideline for IR evaluation}}.
\newblock \bibinfo{journal}{\emph{ACM SIGIR Forum}} \bibinfo{volume}{54},
  \bibinfo{number}{1} (\bibinfo{date}{jun} \bibinfo{year}{2020}),
  \bibinfo{pages}{1--8}.
\newblock
\showISSN{0163-5840}
\urldef\tempurl%
\url{https://doi.org/10.1145/3451964.3451976}
\showDOI{\tempurl}


\bibitem[Sharifzadeh et~al\mbox{.}(2020)]%
        {DBLP:conf/icpr/SharifzadehBBKT20}
\bibfield{author}{\bibinfo{person}{Sahand Sharifzadeh},
  \bibinfo{person}{Sina~Moayed Baharlou}, \bibinfo{person}{Max Berrendorf},
  \bibinfo{person}{Rajat Koner}, {and} \bibinfo{person}{Volker Tresp}.}
  \bibinfo{year}{2020}\natexlab{}.
\newblock \showarticletitle{Improving Visual Relation Detection using Depth
  Maps}. In \bibinfo{booktitle}{\emph{25th International Conference on Pattern
  Recognition, {ICPR} 2020, Virtual Event / Milan, Italy, January 10-15,
  2021}}. \bibinfo{publisher}{{IEEE}}, \bibinfo{pages}{3597--3604}.
\newblock
\urldef\tempurl%
\url{https://doi.org/10.1109/ICPR48806.2021.9412945}
\showDOI{\tempurl}


\bibitem[Sun et~al\mbox{.}(2019)]%
        {sun2018rotate}
\bibfield{author}{\bibinfo{person}{Zhiqing Sun}, \bibinfo{person}{Zhi-Hong
  Deng}, \bibinfo{person}{Jian-Yun Nie}, {and} \bibinfo{person}{Jian Tang}.}
  \bibinfo{year}{2019}\natexlab{}.
\newblock \showarticletitle{RotatE: Knowledge Graph Embedding by Relational
  Rotation in Complex Space}. In \bibinfo{booktitle}{\emph{International
  Conference on Learning Representations}}.
\newblock
\urldef\tempurl%
\url{https://openreview.net/forum?id=HkgEQnRqYQ}
\showURL{%
\tempurl}


\bibitem[Sun et~al\mbox{.}(2020)]%
        {DBLP:journals/pvldb/SunZHWCAL20}
\bibfield{author}{\bibinfo{person}{Zequn Sun}, \bibinfo{person}{Qingheng
  Zhang}, \bibinfo{person}{Wei Hu}, \bibinfo{person}{Chengming Wang},
  \bibinfo{person}{Muhao Chen}, \bibinfo{person}{Farahnaz Akrami}, {and}
  \bibinfo{person}{Chengkai Li}.} \bibinfo{year}{2020}\natexlab{}.
\newblock \showarticletitle{A Benchmarking Study of Embedding-based Entity
  Alignment for Knowledge Graphs}.
\newblock \bibinfo{journal}{\emph{Proc. {VLDB} Endow.}} \bibinfo{volume}{13},
  \bibinfo{number}{11} (\bibinfo{year}{2020}), \bibinfo{pages}{2326--2340}.
\newblock
\urldef\tempurl%
\url{http://www.vldb.org/pvldb/vol13/p2326-sun.pdf}
\showURL{%
\tempurl}


\bibitem[Teru et~al\mbox{.}(2020)]%
        {teru2020}
\bibfield{author}{\bibinfo{person}{Komal~K. Teru}, \bibinfo{person}{Etienne
  Denis}, {and} \bibinfo{person}{Will Hamilton}.}
  \bibinfo{year}{2020}\natexlab{}.
\newblock \showarticletitle{Inductive Relation Prediction by Subgraph
  Reasoning}. In \bibinfo{booktitle}{\emph{Proceedings of the 37th
  International Conference on Machine Learning, {ICML} 2020, 13-18 July 2020,
  Virtual Event}} \emph{(\bibinfo{series}{Proceedings of Machine Learning
  Research}, Vol.~\bibinfo{volume}{119})}. \bibinfo{publisher}{{PMLR}},
  \bibinfo{pages}{9448--9457}.
\newblock
\urldef\tempurl%
\url{http://proceedings.mlr.press/v119/teru20a.html}
\showURL{%
\tempurl}


\bibitem[Tiwari et~al\mbox{.}(2021)]%
        {tiwari2021}
\bibfield{author}{\bibinfo{person}{Sudhanshu Tiwari}, \bibinfo{person}{Iti
  Bansal}, {and} \bibinfo{person}{Carlos~R. Rivero}.}
  \bibinfo{year}{2021}\natexlab{}.
\newblock \showarticletitle{Revisiting the Evaluation Protocol of Knowledge
  Graph Completion Methods for Link Prediction}. In
  \bibinfo{booktitle}{\emph{Proceedings of the Web Conference 2021}}.
  \bibinfo{pages}{809–820}.
\newblock
\urldef\tempurl%
\url{https://doi.org/10.1145/3442381.3449856}
\showDOI{\tempurl}


\bibitem[Toutanova and Chen(2015)]%
        {toutanova2015}
\bibfield{author}{\bibinfo{person}{Kristina Toutanova} {and}
  \bibinfo{person}{Danqi Chen}.} \bibinfo{year}{2015}\natexlab{}.
\newblock \showarticletitle{Observed versus latent features for knowledge base
  and text inference}. In \bibinfo{booktitle}{\emph{Proceedings of the 3rd
  Workshop on Continuous Vector Space Models and their Compositionality}}.
  \bibinfo{publisher}{Association for Computational Linguistics},
  \bibinfo{address}{Beijing, China}, \bibinfo{pages}{57--66}.
\newblock
\urldef\tempurl%
\url{https://doi.org/10.18653/v1/W15-4007}
\showDOI{\tempurl}


\bibitem[Trouillon et~al\mbox{.}(2017)]%
        {DBLP:journals/jmlr/TrouillonDGWRB17}
\bibfield{author}{\bibinfo{person}{Th{\'{e}}o Trouillon},
  \bibinfo{person}{Christopher~R. Dance}, \bibinfo{person}{{\'{E}}ric
  Gaussier}, \bibinfo{person}{Johannes Welbl}, \bibinfo{person}{Sebastian
  Riedel}, {and} \bibinfo{person}{Guillaume Bouchard}.}
  \bibinfo{year}{2017}\natexlab{}.
\newblock \showarticletitle{Knowledge Graph Completion via Complex Tensor
  Factorization}.
\newblock \bibinfo{journal}{\emph{J. Mach. Learn. Res.}}  \bibinfo{volume}{18}
  (\bibinfo{year}{2017}), \bibinfo{pages}{130:1--130:38}.
\newblock


\bibitem[Wang et~al\mbox{.}(2017)]%
        {wang2017}
\bibfield{author}{\bibinfo{person}{Quan Wang}, \bibinfo{person}{Zhendong Mao},
  \bibinfo{person}{Bin Wang}, {and} \bibinfo{person}{Li Guo}.}
  \bibinfo{year}{2017}\natexlab{}.
\newblock \showarticletitle{Knowledge Graph Embedding: A Survey of Approaches
  and Applications}.
\newblock \bibinfo{journal}{\emph{IEEE Transactions on Knowledge and Data
  Engineering}} \bibinfo{volume}{29}, \bibinfo{number}{12}
  (\bibinfo{year}{2017}), \bibinfo{pages}{2724--2743}.
\newblock
\urldef\tempurl%
\url{https://doi.org/10.1109/TKDE.2017.2754499}
\showDOI{\tempurl}


\bibitem[Wang et~al\mbox{.}(2019)]%
        {DBLP:conf/rep4nlp/WangRGBM19}
\bibfield{author}{\bibinfo{person}{Yanjie Wang}, \bibinfo{person}{Daniel
  Ruffinelli}, \bibinfo{person}{Rainer Gemulla}, \bibinfo{person}{Samuel
  Broscheit}, {and} \bibinfo{person}{Christian Meilicke}.}
  \bibinfo{year}{2019}\natexlab{}.
\newblock \showarticletitle{On Evaluating Embedding Models for Knowledge Base
  Completion}. In \bibinfo{booktitle}{\emph{Proceedings of the 4th Workshop on
  Representation Learning for NLP, RepL4NLP@ACL 2019, Florence, Italy, August
  2, 2019}}, \bibfield{editor}{\bibinfo{person}{Isabelle Augenstein},
  \bibinfo{person}{Spandana Gella}, \bibinfo{person}{Sebastian Ruder},
  \bibinfo{person}{Katharina Kann}, \bibinfo{person}{Burcu Can},
  \bibinfo{person}{Johannes Welbl}, \bibinfo{person}{Alexis Conneau},
  \bibinfo{person}{Xiang Ren}, {and} \bibinfo{person}{Marek Rei}} (Eds.).
  \bibinfo{publisher}{Association for Computational Linguistics},
  \bibinfo{pages}{104--112}.
\newblock
\urldef\tempurl%
\url{https://doi.org/10.18653/v1/w19-4313}
\showDOI{\tempurl}


\bibitem[Wu et~al\mbox{.}(2019)]%
        {DBLP:conf/ijcai/WuLF0Y019}
\bibfield{author}{\bibinfo{person}{Yuting Wu}, \bibinfo{person}{Xiao Liu},
  \bibinfo{person}{Yansong Feng}, \bibinfo{person}{Zheng Wang},
  \bibinfo{person}{Rui Yan}, {and} \bibinfo{person}{Dongyan Zhao}.}
  \bibinfo{year}{2019}\natexlab{}.
\newblock \showarticletitle{Relation-Aware Entity Alignment for Heterogeneous
  Knowledge Graphs}. In \bibinfo{booktitle}{\emph{Proceedings of the
  Twenty-Eighth International Joint Conference on Artificial Intelligence,
  {IJCAI} 2019, Macao, China, August 10-16, 2019}},
  \bibfield{editor}{\bibinfo{person}{Sarit Kraus}} (Ed.).
  \bibinfo{publisher}{ijcai.org}, \bibinfo{pages}{5278--5284}.
\newblock
\urldef\tempurl%
\url{https://doi.org/10.24963/ijcai.2019/733}
\showDOI{\tempurl}


\bibitem[Zhu et~al\mbox{.}(2021)]%
        {DBLP:journals/corr/abs-2106-06935}
\bibfield{author}{\bibinfo{person}{Zhaocheng Zhu}, \bibinfo{person}{Zuobai
  Zhang}, \bibinfo{person}{Louis{-}Pascal A.~C. Xhonneux}, {and}
  \bibinfo{person}{Jian Tang}.} \bibinfo{year}{2021}\natexlab{}.
\newblock \showarticletitle{Neural Bellman-Ford Networks: {A} General Graph
  Neural Network Framework for Link Prediction}.
\newblock \bibinfo{journal}{\emph{CoRR}}  \bibinfo{volume}{abs/2106.06935}
  (\bibinfo{year}{2021}).
\newblock
\showeprint[arXiv]{2106.06935}
\urldef\tempurl%
\url{https://arxiv.org/abs/2106.06935}
\showURL{%
\tempurl}


\end{thebibliography}

\appendix
\clearpage

\section{Additional Tables}
\subsection{Generalized Hölder Mean}
\begin{table}[!h]
    \centering
    \caption{%
    Aggregation functions formulated with the generalized Hölder (i.e., power) mean $M_p(x_1,\ldots,x_n)=\sqrt[p]{\frac{1}{n} \sum_{i=1}^n x_i^p}$ as defined in~\cite{Bullen2003}.
    Note that $\lim_{p \rightarrow 0}M_p$ asymptotically approaches the geometric mean.
    }
    \label{tab:aggregations}
    \def\arraystretch{1.2}
    \begin{tabular}{ lrc } 
    \toprule
    \textbf{Name}   & \textbf{p} & \textbf{Definition} \\ 
    \midrule
    \vspace{-2mm}
    Max             & $+\infty$ & $\max \limits_i f(r_i)$ \\
    
    \vdots          & \vdots    & \vdots  \\
    Quadratic Mean  & 2         & $\sqrt[2]{\frac{1}{n} \sum_{i=1}^n f(r_i)^2}$ \\ 
    Arithmetic Mean & 1         & $\frac{1}{n}\sum_{i=1}^n f(r_i)$ \\ 
    Geometric Mean  & 0         & $\sqrt[n]{\prod_{i=1}^n f(r_i)}$ \\  
    Harmonic Mean   & -1        & $\big(\frac{1}{n}\sum_{i=1}^n f(r_i)^{-1}\big)^{-1}$ \\
    \vspace{-1mm}
    \vdots          & \vdots    &  \vdots \\
    Min             & $-\infty$ & $\min \limits_i f(r_i)$ \\
    \bottomrule
    \end{tabular}
\end{table}

\subsection{Datasets}

As standard rank-based metrics depend on the number of entities, we chose datasets whose number of entities spanned several orders of magnitudes, from $10^1$ to $10^4$, for the case study presented in \autoref{sec:case-study}.
We present statistics on these datasets in \autoref{tab:datasets}. 

\begin{table}[!h]
    \small
    \centering
    \def\arraystretch{1.2}
    \caption{Dataset statistics}
    \label{tab:datasets}
    \begin{tabular}{ lrrr } 
    \toprule
    \textbf{Dataset}   & $|\mathcal{E}|$  & $|\mathcal{R}|$  & $|\mathcal{T}_{\text{train}}|$    \\ 
    \midrule
    Nations      & 14                       & 55  & 1,592  \\
    Kinships         & 104  & 25 & 8,544 \\
    FB15k-237   & 14,505 & 237 & 272,115 \\ 
    WN18-RR & 40,559 & 11 & 86,835 \\
    \bottomrule
    \end{tabular}
\end{table}

\section{Derivations of Adjustments}\label{sec:adjustements}

For derivation of the adjustments, we assume each ranking task $r_i$ to be independent and identically distributed (i.i.d.) according to a discrete uniform distribution $r_i \sim \mathcal{U}(1, N_i) \in [1,\ldots,N_i]$.
While the upper bound $N_i$ \textit{may} vary by ranking task $i$, e.g., due to filtered evaluation, we also provide simplified formulas for the case it remains constant throughout the following derivations such that $\forall i: N_i = N$.
We denote equivalences asserted under this assumption with $\stackrel{*}{=}$.

\subsection{Adjusting the \ac{MR}}

We begin by briefly recapitulating the derivation of the adjusted (arithmetic) mean rank from~\cite{berrendorf2020} by first deriving the expectation of the \ac{MR} (\autoref{eq:mr-expectation}).
The expectation and variance of a uniformly distributed discrete variable  $X \sim \mathcal{U}(a, b)$ are respectively $\mathbb{E}\left[X\right] = \frac{b+a}{2}$ and $\Var\left[X\right] = \frac{\left(b-a+1\right)^2-1}{12}$.
Given our uniformly distributed variable $r_i$ with parameters $a=1$ and $b=N_i$, we get the following expectation:

\begin{equation} \label{eq:mr-func-expectation}
\mathbb{E}\left[r_i\right] = \frac{N_i+1}{2} \stackrel{*}{=} \frac{N+1}{2}
\end{equation}

\noindent
The variance of $r_i$ is given as:

\begin{equation} \label{eq:mr-func-variance}
\Var\left[r_i\right]
= \frac{\left(N_i+1-1\right)^2-1}{12}
\stackrel{*}{=} \frac{N^2-1}{12}
\end{equation}

\noindent
Consequently, the expectation of the \ac{MR} metric is given as:

\begin{equation} \label{eq:mr-expectation}
\mathbb{E}\left[\textrm{MR}\right]
= \mathbb{E}\left[\frac{1}{n} \sum \limits_{i=1}^n r_i\right]
= \frac{1}{n} \sum \limits_{i=1}^n \mathbb{E}\left[ r_i\right]
= \frac{1}{n} \sum \limits_{i=1}^n \frac{N_i+1}{2}
\stackrel{*}{=} \frac{N+1}{2}
\end{equation}

\noindent
The variance of the \ac{MR} metric is given as:

\begin{equation} \label{eq:mr-metric-variance}
\begin{split}
\Var\left[\textrm{MR}\right]
& = \Var\left[\frac{1}{n} \sum \limits_{i=1}^n r_i\right]
= \frac{1}{n} \sum \limits_{i=1}^n \Var\left[ r_i\right] \\
& = \frac{1}{n} \sum \limits_{i=1}^n \frac{N_i^2-1}{12}
\stackrel{*}{=} \frac{N^2-1}{12}
\end{split}
\end{equation}

\subsubsection{Chance-adjusted \ac{MR}}

The chance-adjusted \ac{MR} (called \emph{adjusted mean rank (AMR)} in \cite{berrendorf2020}) is given as:

\begin{equation} \label{eq:mr-adjusted-definition}
\textrm{MR}^{*}(r_1,\ldots,r_n)
= \frac{\textrm{MR}(r_1,\ldots,r_n)}{\mathbb{E}\left[\textrm{MR}\right] }
\stackrel{*}{=} \frac{2}{N(N-1)}\sum_{i=1}^n r_i
\end{equation}

\subsubsection{Re-indexed Chance-adjusted \ac{MR}}

The authors of \cite{berrendorf2020} introduced a re-indexed variant of the \ac{AMR} named \ac{AMRI} that is given as follows:

\begin{equation} \label{eq:amri-definition}
\textrm{AMRI}(r_1,\ldots,r_n)
= 1 - \frac{MR(r_1,\ldots,r_n) - 1}{\mathbb{E}\left[\text{MR}-1\right]} \hspace{5mm} \in [-1,1]
\end{equation}

\subsection{Adjusting the \ac{MRR}}

The expectation and variance of an inverse-uniform distributed variable\footnote{\url{https://en.wikipedia.org/wiki/Inverse_distribution\#Inverse_uniform_distribution}} $\frac{1}{X} \sim \mathcal{U}\left(\frac{1}{a},\frac{1}{b}\right)$ are $\mathbb{E}\left[\frac{1}{X}\right] = \frac{\ln b - \ln a}{b - a}$ and $\Var\left[\frac{1}{X}\right]=\frac{1}{ab}-\big(\frac{\ln b - \ln a}{b-a}\big)^2$.
Given our uniformly distributed variable $r_i$  with parameters $a=1$ and $b=N_i$ and its corresponding inverse-uniform distributed variable $r_i^{-1}$, we get the following expectation:

\begin{equation} \label{eq:inverse-uniform-expectation}
\mathbb{E}\left[r_i^{-1}\right]
= \frac{\ln 1 - \ln N_i}{N_i - 1}
= \frac{\ln N_i}{N_i - 1}
\stackrel{*}{=} \frac{\ln N}{N - 1}
\end{equation}

\noindent
The variance of $r_i$ is given as:

\begin{equation} \label{eq:inverse-uniform-variance}
\Var\left[r_i^{-1}\right]
= \frac{1}{1 \cdot N_i} - \left(\frac{\ln N_i - \ln 1}{N_i - 1}\right)
\stackrel{*}{=} \frac{1}{N} - \frac{\ln N}{N - 1}
\end{equation}

\noindent
The expectation of the \ac{MRR} metric is given as:

\begin{equation} \label{eq:mrr-expectation}
\mathbb{E}\left[\textrm{MRR}\right] 
= \mathbb{E}\left[\frac{1}{n} \sum \limits_{i=1}^n r_i^{-1}\right] 
= \frac{1}{n} \sum \limits_{i=1}^n \mathbb{E}\left[r_i^{-1}\right] 
= \mathbb{E}\left[r_i^{-1}\right]
\stackrel{*}{=} \frac{\ln N}{N - 1}
\end{equation}

\noindent
The variance of the \ac{MRR} metric is given as:

\begin{equation} \label{eq:mrr-metric-variance}
\begin{split}
\Var\left[\textrm{MRR}\right]
& = \Var\left[\frac{1}{n} \sum \limits_{i=1}^n r_i^{-1}\right]
= \frac{1}{n} \sum \limits_{i=1}^n \Var\left[ r_i^{-1}\right] \\
& = \frac{1}{n} \sum \limits_{i=1}^n \frac{1}{N_i} - \frac{\ln N_i}{N_i-1}
\stackrel{*}{=} \frac{1}{N} - \frac{\ln N}{N-1}
\end{split}
\end{equation}

\subsubsection{Chance-adjusted \ac{MRR}}

The chance-adjusted \ac{MRR} is given as:

\begin{equation} \label{eq:mrr-adjusted-definition}
\textrm{MRR}^{*}(r_1,\ldots,r_n)
= \frac{\textrm{MRR}(r_1,\ldots,r_n)}{\mathbb{E}\left[\textrm{MRR}\right] }
\stackrel{*}{=} \frac{N-1}{N \ln N}\sum_{i=1}^n r_i^{-1}
\end{equation}

\subsection{Adjusting the \Acl{HK}}

The expectation of \ac{HK} is derived first by deriving the expectation of the discrete indicator function $f(x) = \mathbb{I}\left[ x \le k \right]$ (\autoref{eq:discrete-indicator-expectation}) then applying it in full under the assumption that $N_i = N$ for all $i$ (\autoref{eq:hk-expectation}). The expectation of $r_i$ is given as:

\begin{equation} \label{eq:discrete-indicator-expectation}
\mathbb{E}\left[\mathbb{I}[r_i \leq k]\right] = \frac{k}{N_i} \stackrel{*}{=} \frac{k}{N}
\end{equation}

\noindent
The variance of $r_i$ is given as:

\begin{equation} \label{eq:discrete-indicator-variance}
\begin{split}
\Var\left[\mathbb{I}[r_i \leq k]\right] 
& = \mathbb{E}\left[\mathbb{I}[r_i \leq k]\right] \times (1 - \mathbb{E}\left[\mathbb{I}[r_i \leq k]\right]) \\ 
& = \frac{k}{N_i} \times \left(1 - \frac{k}{N_i}\right) = \frac{k(N_i-k)}{N_i} \\
& \stackrel{*}{=} \frac{k}{N} \times \left(1 - \frac{k}{N}\right) = \frac{k(N-k)}{N}
\end{split}
\end{equation}

\noindent
The expectation of the \ac{HK} metric is given as:

\begin{equation} \label{eq:hk-expectation}
\begin{split}
\mathbb{E}[H_k] 
&= \mathbb{E}\left[\frac{1}{n} \sum \limits_{i=1}^n \mathbb{I}[r_i \leq k]
\right]
= \frac{1}{n} \sum \limits_{i=1}^n \mathbb{E}\left[\mathbb{I}[r_i \leq k]\right] \\
& = \frac{1}{n} \sum \limits_{i=1}^n \frac{k}{N_i}
\stackrel{*}{=}\frac{k}{N}
\end{split}
\end{equation}

The variance of the \ac{HK} metric is given as:

\begin{equation} \label{eq:hk-variance}
\begin{split}
\Var[H_k] 
&= \Var\left[\frac{1}{n} \sum \limits_{i=1}^n \mathbb{I}[r_i \leq k]
\right]
= \frac{1}{n} \sum \limits_{i=1}^n \Var\left[\mathbb{I}[r_i \leq k]\right] \\
& = \frac{1}{n} \sum \limits_{i=1}^n \frac{k(N_i-k)}{N_i}
\stackrel{*}{=}\frac{k(N-k)}{N}
\end{split}
\end{equation}

\subsubsection{Chance-adjusted \ac{HK}}

The chance-adjusted \ac{HK} is given as:

\begin{equation} \label{eq:hk-adjusted-definition}
H_k^{*}(r_1,\ldots,r_n)
= \frac{H_k(r_1,\ldots,r_n)}{\mathbb{E}\left[H_k\right] }
= \frac{1}{k}\sum_{i=1}^n \mathbb{I}[r_i \leq k]
\end{equation}

\subsubsection{Re-indexed Chance-adjusted \ac{HK}}

Combining the facts that ranks are 1-indexed and the $H_k\in[0,1]$, the \ac{HK} can be adjusted as in \autoref{eq:hk-reindexed-adjusted-definition}.
A negative value of the \ac{AHK} corresponds to performance below random, zero corresponds to random performance, and 1 to optimal performance.
The adjustment for \ac{HK} is affine with respect to a dataset's filtering constant, so it can be applied to results \textit{after} evaluation.

\begin{equation} \label{eq:hk-reindexed-adjusted-definition}
AH@k = \frac{H_k - \mathbb{E}[H_k]}{1 - \mathbb{E}[H_k]} \hspace{5mm} \in \left(-\frac{\mathbb{E}[H_k]}{1-\mathbb{E}[H_k]}, 1\right]
\end{equation}

\noindent
Note the lower bound was calculated by inserting $\min H_k$ as the value for $H_k$, which is 0.

\subsection{Remaining Adjustments}

Identifying a closed-form expectation for the \acf{GMR} is difficulty because of the inclusion of a product.
Further, identifying closed-form expectations for \acf{HMR}, \acf{IGMR}, and \acf{IMR} come from the difficulty of introducing inverses.
For these, we implemented a simple workflow to numerically estimate the adjustment constant in PyKEEN that can be applied as an affine transformation after the fact.

\end{document}